# Illumination-Invariant Anomaly Detection for Sub-Canopy UAV Multispectral Point Clouds

Likun Chen
School of Electronics and Information Engineering
Harbin Institute of Technology
Harbin, China
20b905008@stu.hit.edu.cn

Yanfeng Gu
School of Electronics and Information Engineering
Harbin Institute of Technology
Harbin, China
guyf@hit.edu.cn

Xian Li*
School of Electronics and Information Engineering
Harbin Institute of Technology
Harbin, China
xianli@hit.edu.cn

***Abstract*—Unmanned Aerial Vehicle (UAV) multispectral point clouds (MPC) provide high-dimensional spatial-spectral data for sub-canopy target detection; however, their efficacy is significantly compromised by severe illumination heterogeneity caused by vegetation shadows. To address this, we propose a prior-free anomaly detection framework capable of robustly handling lighting variations. First, we formulate solar angle estimation as an inverse optimization problem. By coupling spectral indices with a ray-tracing model, this strategy achieves Prior-Free Shadow Extraction without relying on flight metadata, effectively distinguishing dark objects from true shadows. Second, to mitigate spectral distortions, we introduce an Illumination-Consistent Sparse Representation mechanism. Unlike standard reconstruction methods, we construct a background dictionary strictly from neighbors sharing the same illumination state. This constraint effectively disentangles material reflectance from lighting variations, ensuring that targets are represented solely by physically consistent background points. Experimental results indicate that the proposed method significantly improves the separability between anomalies and background in complex forest environments, demonstrating superior performance over state-of-the-art baselines. This framework is particularly suited for identifying camouflaged military targets, mapping fallen tree trunks, and uncovering archaeological ruins hidden beneath dense foliage.**

## I. INTRODUCTION

UAV multispectral point clouds provide high-resolution 4D spatial-spectral representations for sub-canopy target detection [1], [2]. Nevertheless, in complex forests, vegetation shadows induce severe illumination heterogeneity which degrades spectral separability and signal-to-noise ratios [3]. Detecting sub-canopy anomalies is of paramount practical importance, specifically for tasks such as locating camouflaged military assets, identifying fallen trunks in forest management, and uncovering archaeological vestiges [4]. Current anomaly detection methodologies broadly fall into data-driven learning and statistical modeling. Deep neural networks excel in object processing but inherently rely on extensive training samples, limiting their utility in unsupervised scenarios lacking target priors. Conversely, statistical baselines derived from hyperspectral imaging like the Reed-Xiaoli detector [4] enable detection without training samples by modeling background distributions. Nevertheless, adapting these grid-based algorithms to unstructured point clouds is hindered by variable density and anisotropy which complicate neighborhood definitions. Although recent spherical shell models successfully address spatial isolation, they prioritize statistical purity over spectral consistency. Specifically, mechanisms to compensate for illumination-induced spectral distortion remain underexplored, leaving shadow interference in sub-canopy environments as a critical challenge requiring physically constrained investigation [5].

Existing solutions for handling these shadows face distinct limitations regarding extraction accuracy and integration strategies. First, deep learning approaches suffer from poor generalization when scene characteristics shift [6]. Simultaneously, traditional spectral indices relying on thresholding frequently misclassify dark materials like water or asphalt as shadows due to spectral ambiguity. Furthermore, physics-based ray-tracing models require precise solar azimuth and elevation parameters, which are often missing from standard UAV metadata. In terms of data dependency, current imagery-based methods often combine data with Digital Surface Models (DSM) and Digital Elevation Models (DEM) [7]. However, reliance on such auxiliary elevation data often introduces issues related to low spatial resolution and reconstruction errors. Finally, regarding the integration framework, the predominant sequential restoration strategy involving shadow removal followed by detection is susceptible to error propagation where artifacts introduced during compensation accumulate and impair precision. Consequently, an integrated framework operating independently of environmental priors is essential to address shadow interference in unstructured MPC data.

To address the limitations of relying on external priors and the error propagation inherent in sequential strategies, this paper proposes an unsupervised, illumination-invariant anomaly detection framework for UAV MPCs. Unlike conventional approaches that rely on external parameters or sequential processing, we develop a strategy grounded in physical constraints. We first introduce an iterative method for shadow estimation that couples spectral indices with a ray tracing model based on physics. This approach leverages the precise native geometry of LiDAR to differentiate precisely between points in sunlight and those in shadow without requiring external solar priors. Guided by this accurate distribution, we employ a Spherical Shell Model (SSM) to construct background dictionaries that adapt to local illumination conditions. By reconstructing spatial and spectral features according to their specific lighting environment, the representation error serves directly as a robust anomaly score. This method effectively mitigates the impact of heterogeneous illumination and achieves robust anomaly detection without training samples.

## II. METHODOLOGY

The proposed framework comprises two stages: Prior-Free Shadow Extraction and Illumination-Invariant Anomaly Detection. The overall workflow is illustrated in Fig. 1.

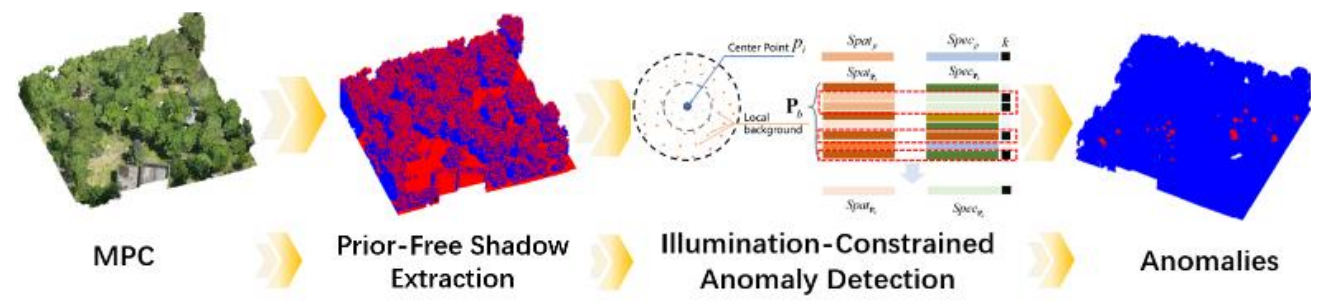

Fig. 1. Workflow for sub-canopy anomaly detection of MPC.

### A. Prior-Free Shadow Extraction via MPC

Accurate shadow extraction is critical for decoupling illumination effects but is often hindered by the misclassification of dark materials (e.g., water, asphalt) as shadows due to their low reflectance. To address this without prior solar information, we propose a coarse-to-fine strategy that synergizes spectral analysis with geometric verification.

Initially, a coarse shadow mask is derived using the Normalized Saturation-Value Difference Index (NSVDI), defined as $NSVDI = (V - S)/(V + S)$, where $V$ represents the brightness value, calculated as the maximum reflectance across all multispectral bands, and $S$ denotes the spectral saturation or variation. In sub-canopy environments, shadows exhibit a significant decrease in $V$ due to direct light occlusion, while $S$ relatively increases because of sky diffuse radiation and, more importantly, the selective transmission and multiple scattering of light through the vegetation canopy. This low-brightness and high-saturation characteristic provides a robust physical basis for distinguishing true shadows from dark materials.

The MPC data used in this study covers a wide spectral range from visible to near-infrared light, comprising 10 bands with center wavelengths of 444, 475, 531, 560, 650, 668, 705, 717, 740, and 842 nm. This rich spectral information supports the application of mature spectral indices for rapid shadow extraction. We conducted a comparative evaluation of representative methods including spectral indices, PCA, and color space transformations. Experimental results demonstrate that NSVDI outperforms comparative methods in shadow extraction tasks. Although NSVDI may generate minor pixel-level misclassifications due to spectral ambiguity, it accurately captures the dominant global distribution of shadows. This provides a robust initialization for the subsequent inverse optimization, ensuring that the physics-based ray-tracing model converges to the global optimum rather than getting stuck in local minima..

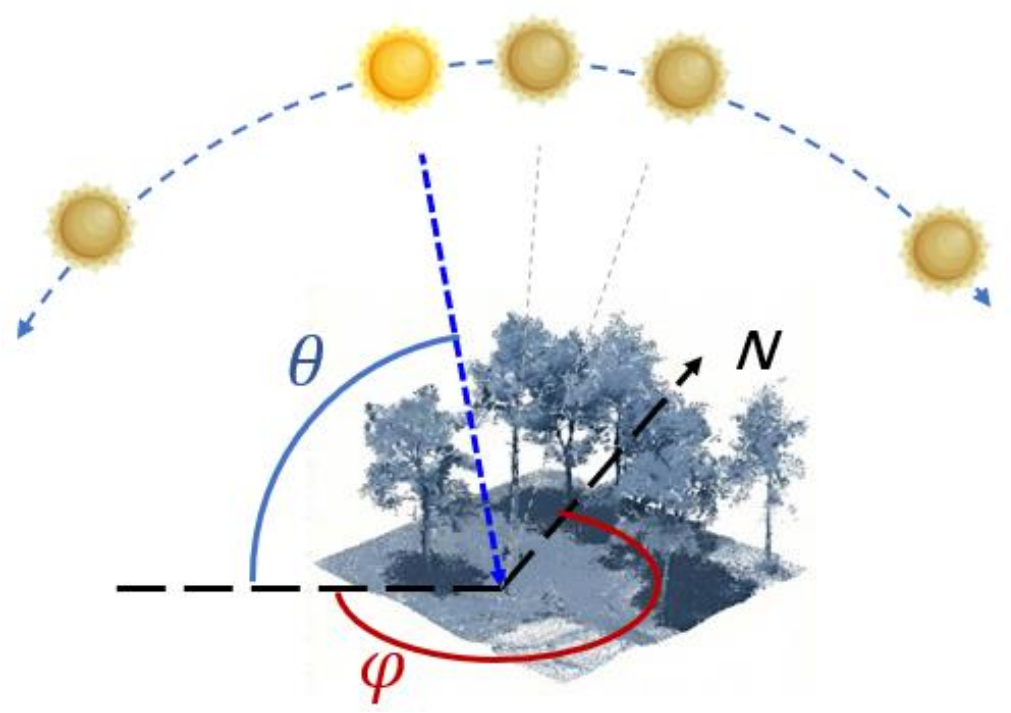

Fig. 2. Iteration of solar altitude angle $\theta$ and azimuth angle $\varphi$ of MPC

Second, to rectify false positives caused by spectral ambiguities, we introduce a geometric constraint based on a Ray-Tracing model. Given that precise solar elevation $\theta$ and azimuth $\varphi$ are frequently omitted from metadata, solar position estimation is formulated as an inverse optimization problem. The stability of this solar angle inversion is maintained through the joint utilization of spectral features and geometric constraints. By iterating through the possible range of solar angles, we generate simulated geometric shadows. The optimal solar angles $(\theta^*, \varphi^*)$ are determined by maximizing the Intersection over Union (IoU) between the coarse spectral mask and the simulated geometric shadow, $(\theta^*, \phi^*) = \arg\max_{\theta,\phi} \mathrm{IoU}(\theta, \phi)$. The geometric projection under these optimal angles yields the final refined illumination factor $k_i$ for each point $p_i$. As detailed in Table I, the final $k_i$ is derived by prioritizing geometric verification: points identified as shadows by spectral indices but confirmed as sunlit by geometric projection are corrected to sunlit status ($k_i = 1$), and $k_i = 0$ denotes shadowed areas. This step effectively separates true shadows from spectral anomalies.

TABLE I. CORRECTION FOR SPECTRAL-GEOMETRIC CONFLICTS

| Land Cover | Spectral | Geometric | Final Label ( $k$ ) |
|---|---|---|---|
| Dark Objects (sunlit) | shadow | sunlit | 1 (sunlit) |
| Water (sunlit) | shadow | sunlit | 1 (sunlit) |
| Dark Objects (shadow) | shadow | shadow | 0 (shadow) |
| Bright Objects (shadow) | sunlit | shadow | 0 (shadow) |

### B. Illumination-Invariant Anomaly Detection

Guided by the estimated illumination distribution, the second stage performs Illumination-Invariant Joint Reconstruction. Unlike linear models that struggle with heterogeneous lighting, we unify spatial and spectral domains under a strict physical consistency constraint.

1) Joint Spatial-Spectral Representation: To rigorously characterize the point attributes, we formulate each point $p_i$ as a comprehensive joint feature vector $\mathbf{y}_i$. First, to capture the local geometric topology, we extract a set of spatial descriptors $\mathbf{y}_i^{\text{spat}}$ derived from the local covariance matrix, specifically including Roughness, Information Entropy, Planarity, Surface Variation, Verticality, Omnivariance, and the First Principal Component. Second, the multispectral intensity values constitute the spectral component $\mathbf{y}_i^{\text{spec}}$. Finally, the estimated illumination factor $k_i$ is explicitly included to define the lighting state. The resulting joint vector is defined as the direct concatenation of these components:

$$\mathbf{y}_i = \left[ (\mathbf{y}_i^{\text{spat}})^{\mathrm{T}}, (\mathbf{y}_i^{\text{spec}})^{\mathrm{T}}, k_i \right]^{\mathrm{T}} \quad (1)$$

This representation ensures that the reconstruction process considers the full context of geometry, material, and illumination.

2) Illumination- Invariant Dictionary Construction: To robustly characterize the complex geometric context of forests, we adopt the Spherical Shell Model (SSM) as the baseline for neighborhood selection. While previous studies have demonstrated the effectiveness of SSM in capturing multi-scale spatial features, standard implementations typically neglect the severe spectral distortion caused by shadows. To address this limitation, we enforce a strict illumination consistency constraint. For a test point $p_i$, its adaptive local background set $P_b$ is constructed by filtering the SSM-derived neighborhood $\mathcal{N}_i$. A neighboring point $p_j$ is selected as a valid dictionary atom if and only if it shares the same illumination label as the center point:

$$P_b = \{ p_j \in \mathcal{N}(i) |\ k_j = k_i \} \quad (2)$$

Consequently, the dictionary $\mathbf{D}_i$ is assembled strictly from these physically consistent neighbors. This constraint effectively disentangles intrinsic material reflectance from extrinsic illumination variations, ensuring that shadowed targets are reconstructed exclusively by background atoms sharing consistent lighting conditions.

3) Joint Reconstruction and Anomaly Scoring: With the constrained dictionary $\mathbf{D}_i$ , we solve for a unified sparse coefficient vector $\boldsymbol{\alpha}$ that minimizes the reconstruction error of the joint feature vector $\mathbf{y}_i$ :

$$\boldsymbol{\alpha}^* = \arg\min_{\boldsymbol{\alpha}} \| \mathbf{y}_i - \mathbf{D}_i\boldsymbol{\alpha}\|_2^2 + \lambda\| \boldsymbol{\alpha}\|_2^2 \tag{3}$$

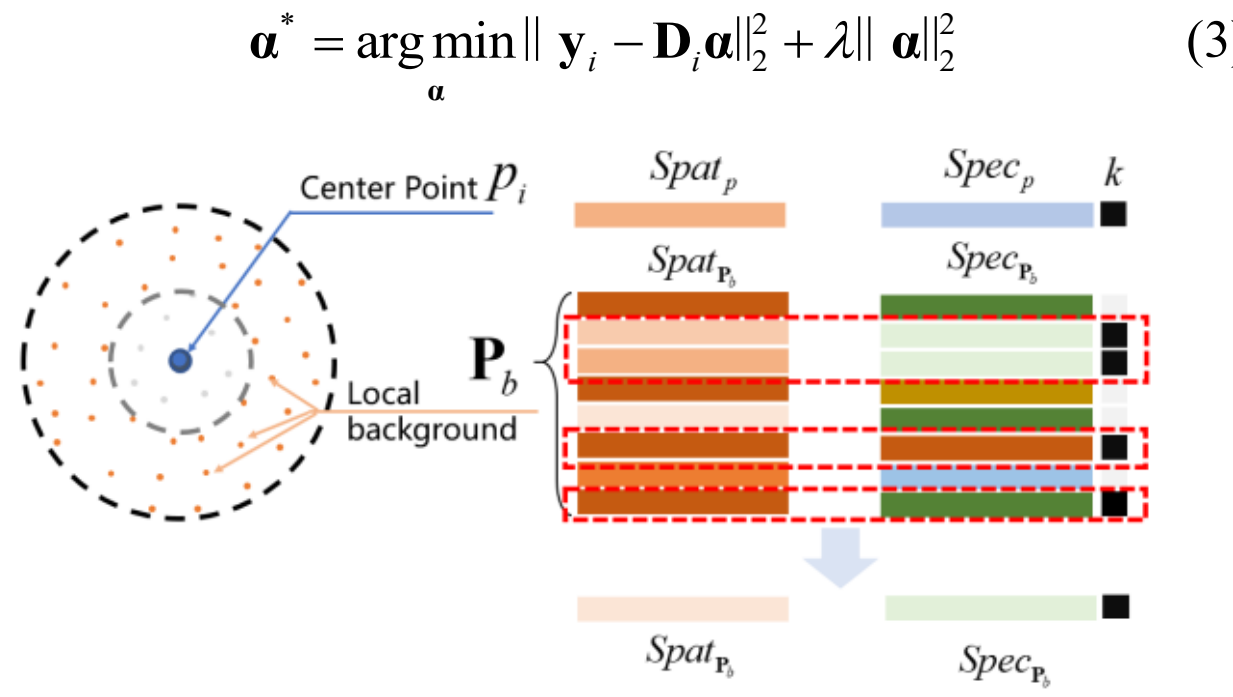

Fig. 3. Illumination-invariant anomaly detection method of MPC.

The final anomaly score $S(p_i)$ is defined as the reconstruction residual. By enforcing illumination consistency, normal background points yield low residuals due to the homogeneity of the dictionary, whereas anomalies exhibit high residuals as they cannot be sparsely represented even by their illumination-consistent neighbors.

## III. Experimental Results and Analysis

### *A. Experiment Equipment and Data*

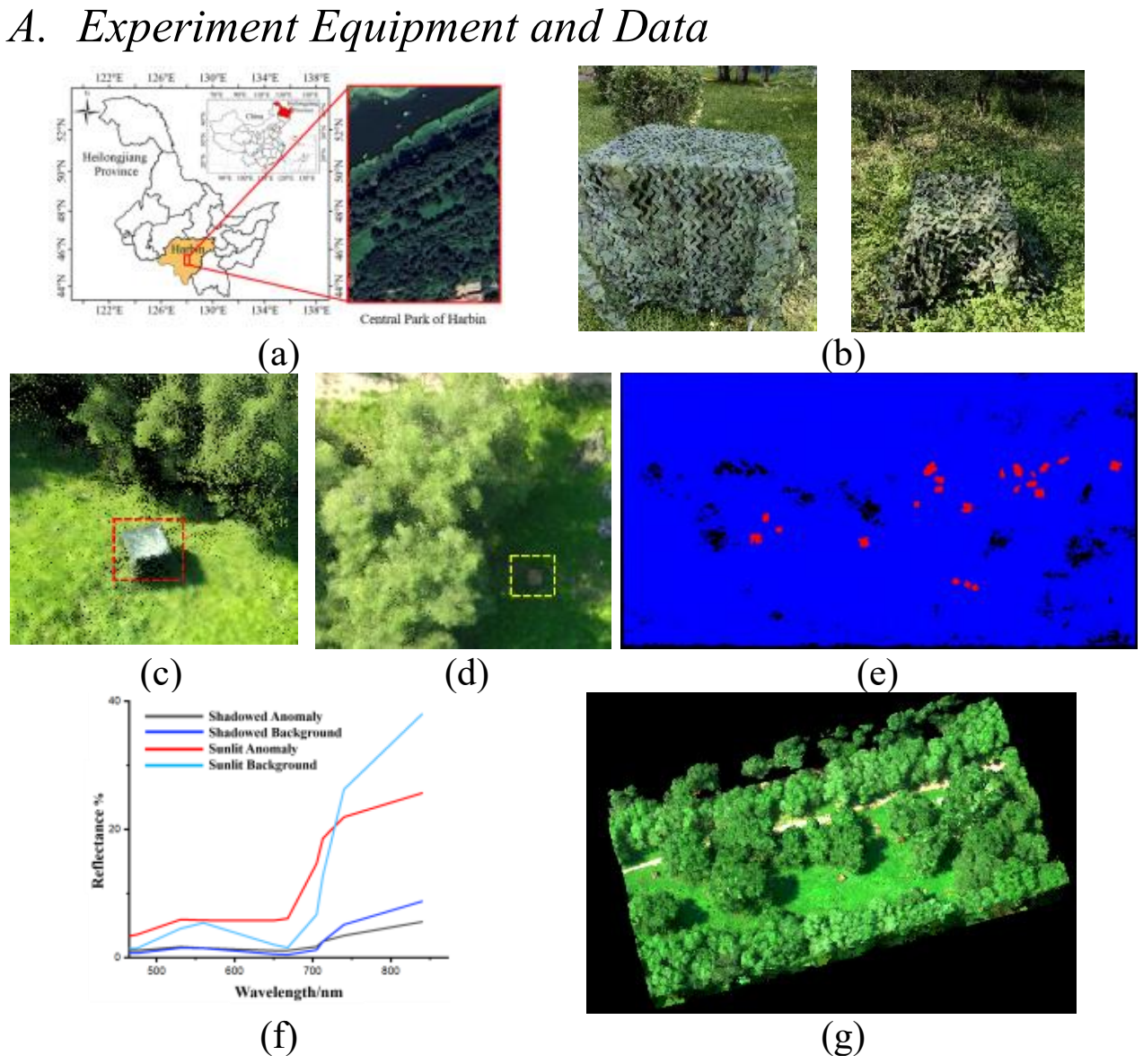

Fig. 4. (a) Dataset acquisition area at Central Park (2024). (b) Anomalies. (c)Sunlit anomaly. (d)Shadowed anomaly. (e) The ground truth.(f) Illustration of spectral ambiguity in shadows. (g) Visualization of the MPC dataset.

The proposed method was validated using a large-scale MPC dataset acquired via a UAV platform over Central Park in Heilongjiang Province, China on June 28, 2024, with a solar elevation angle of 30.86° and an azimuth of 272.65°. To simulate anomalies, cubes covered with camouflage netting with side lengths of 1 m and 0.5 m, as well as flat green fabric panels were positioned within the study area. The flight altitude was 80 m, and the flight speed was 8 m/s. The MPC dataset contains 10-bands captured by the Red MX multispectral camera and the DJI L2 LiDAR, using Agisoft Metashape. The scene includes elm trees, maple trees, willow trees, bald cypress trees, shrubs, grassland, and bare soil roads. The fused MPC contains 3,223,004 points.

### *B. Result and Analysis*

First, the proposed Prior-Free Shadow Extraction is validated on two shadow-rich sub-regions against spectral baselines using solar-angle-derived ground truth as detailed in Fig.5 and Table II. Subsequently, we evaluate the anomaly detection accuracy against the unimproved D3SCRD and four established algorithms: RXD[8], FEBPAD[9], RGAE[10], DUWC[11], and D3SCRD[12]. To ensure a fair comparison, these baselines are pre-processed via Cumulative Distribution Function-based shadow compensation and are denoted as D-RXD, D-FEBPAD, D-RGAE, and D-DUWC, respectively. Qualitative comparisons appear in Fig. 6, while Fig. 7 specifically highlights error distributions at the probability of false alarm $P_{\mathrm{F}} = 0.01$ . This value is selected because $P_{\mathrm{F}} = 0.01$ is a widely used benchmark in anomaly detection applications for evaluating the detection probability, providing a balanced assessment of detection sensitivity and background suppression. Quantitative evaluations are provided in Fig. 8.

Quantitative results in Table II demonstrate that the proposed Prior-Free Shadow Extraction method outperforms traditional spectral index-based approaches in two distinct regions with an accuracy improvement of up to 8%. Notably, the integration with NSVDI yields the highest accuracy of 0.9713.

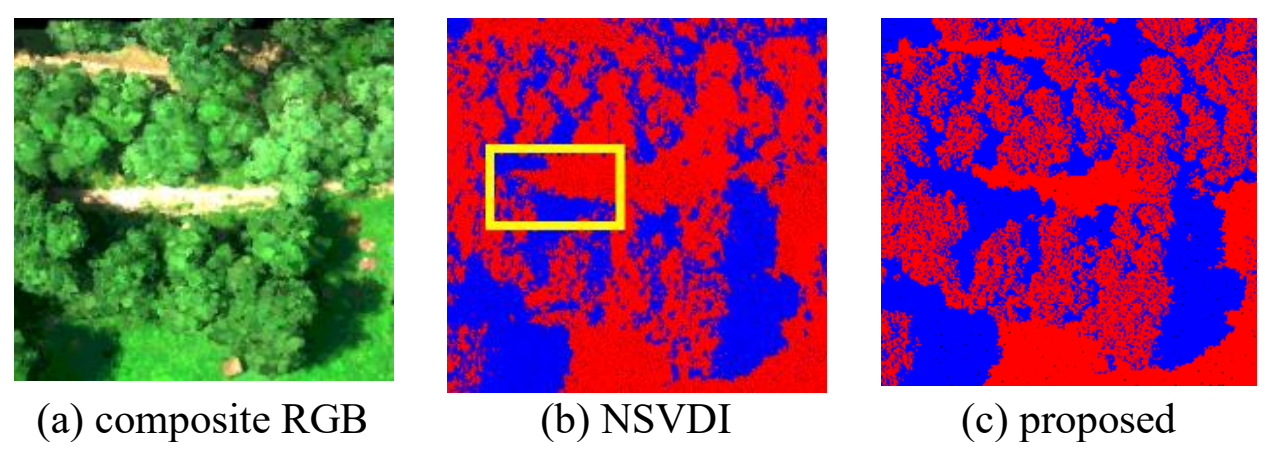

(a) composite RGB (b) NSVDI (c) proposed

Fig. 5. Shadow extraction error of spectral indices.

TABLE II. The Accuracy of Rapid Detection of Shadow Points

| Dataset | Dataset part1 | | Dataset part2 | |
|---|---|---|---|---|
| Method | Original | Proposed | Original | Proposed |
| CIELab | 0.8511 | 0.9090 | 0.8750 | 0.9416 |
| HSV | 0.8522 | 0.9148 | 0.8751 | 0.9345 |
| ISI | 0.8682 | 0.9405 | 0.9034 | 0.948 |
| LSRI | 0.8476 | 0.9315 | 0.8834 | 0.9470 |
| MSDI | 0.8542 | 0.9325 | 0.8778 | 0.9114 |
| NDSI | 0.8710 | 0.9104 | 0.9100 | 0.9241 |
| NSI | 0.7952 | 0.8580 | 0.7819 | 0.8424 |
| SI | 0.8588 | 0.9175 | 0.8847 | 0.9014 |
| SRI | 0.8705 | 0.9465 | 0.8920 | 0.8945 |
| NSVDI | 0.8913 | **0.9713** | 0.9260 | **0.9614** |

For the qualitative evaluation in Figs. 6 and 7, the vegetation canopy is removed to retain only the sub-canopy point cloud where black regions represent sparse background density. Fig. 6 reveals that the proposed method exhibits high response intensity for anomalies within shadows while effectively suppressing background interference such as bare soil and roads. Furthermore, the binary segmentation results at $P_{\mathrm{F}} = 0.01$ in Fig. 7 show that three anomalies located in the left shadowed region are clearly detected with significantly fewer false alarms compared to the unimproved baseline. These results confirm the robustness of the Illumination-Invariant Anomaly Detection method in identifying anomalies within shadowed multispectral point clouds. The ROC curves in Fig. 8(a) show that the proposed algorithm has the highest AUC value associated with the ROC curve, indicating the best detection accuracy, followed by D3SCRD. Combined with the results in Table III, the AUC value of the proposed algorithm reaches 0.9326. In Fig. 8(b), to more clearly illustrate the separation between targets and background, the y-axis is displayed on a logarithmic scale, with the box range from 10% to 90%. The proposed method significantly enhances the separability of anomalies from the background.

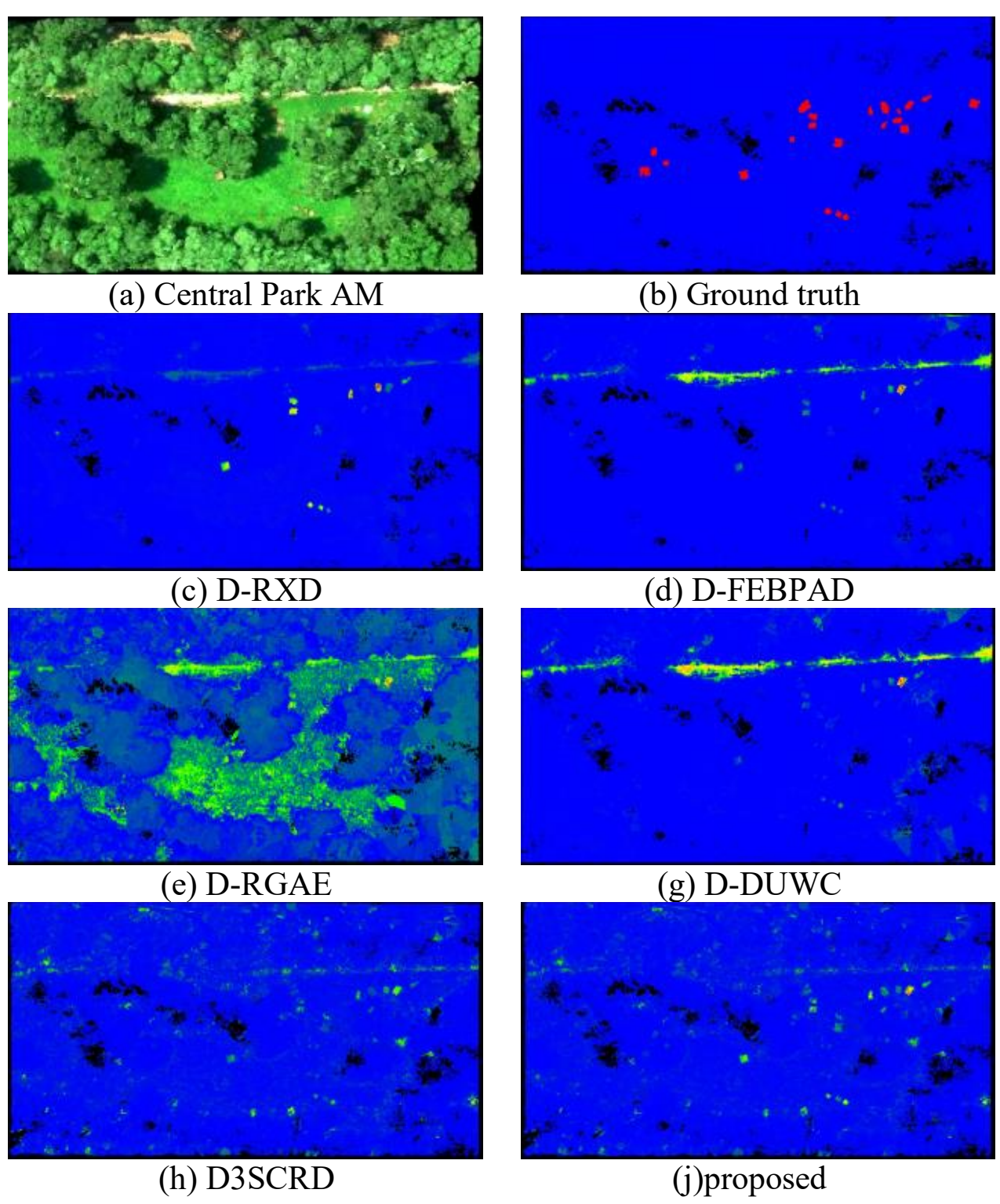

Fig. 6. Anomaly detection results of Central Park 2024 dataset.

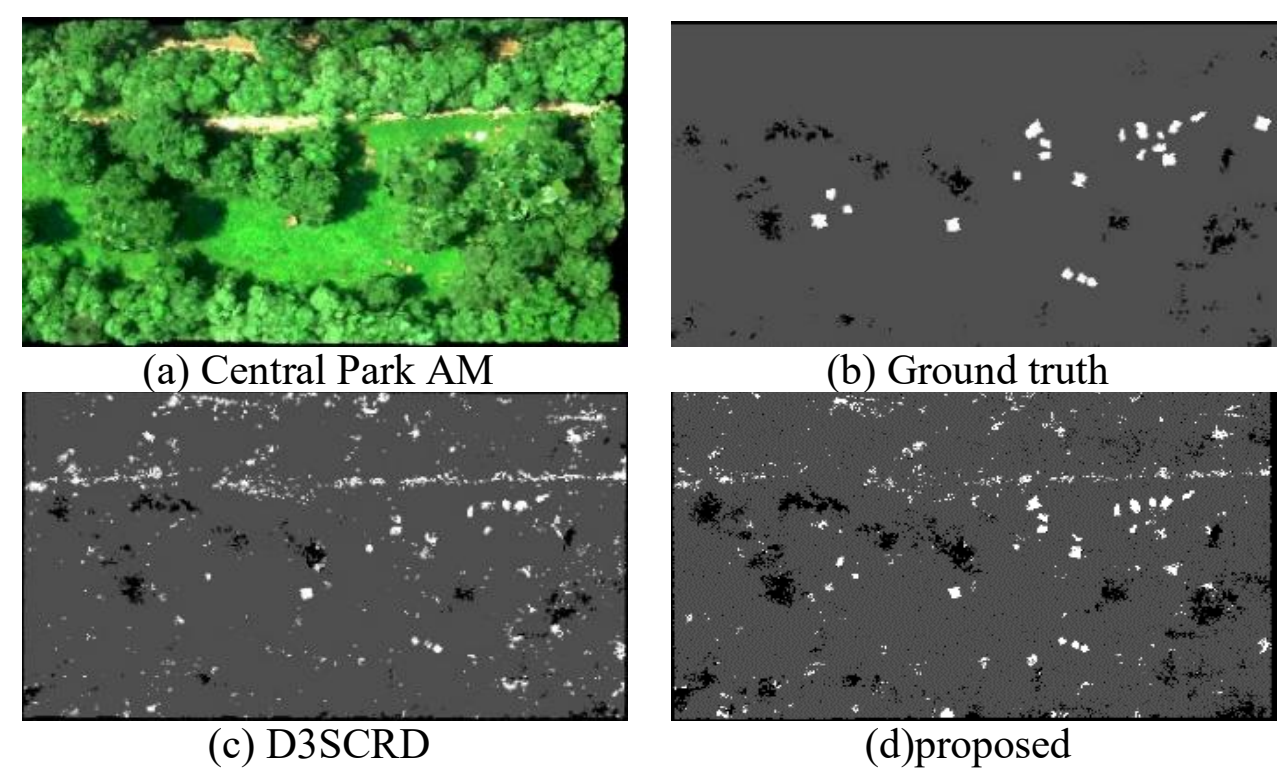

Fig. 7. Binary map of results ( $P_F = 0.01$ ) of Central Park 2024 dataset.

TABLE III. AUC SCORS OF THE ROC CURVES

| Method | AUC |
|---|---|
| D-RXD | 0.8896 |
| D-FEBPAD | 0.8880 |
| D-RGAE | 0.6451 |
| D-DUWC | 0.8153 |
| D3SCRD | <u>0.9188</u> |
| **Proposed** | **0.9326** |

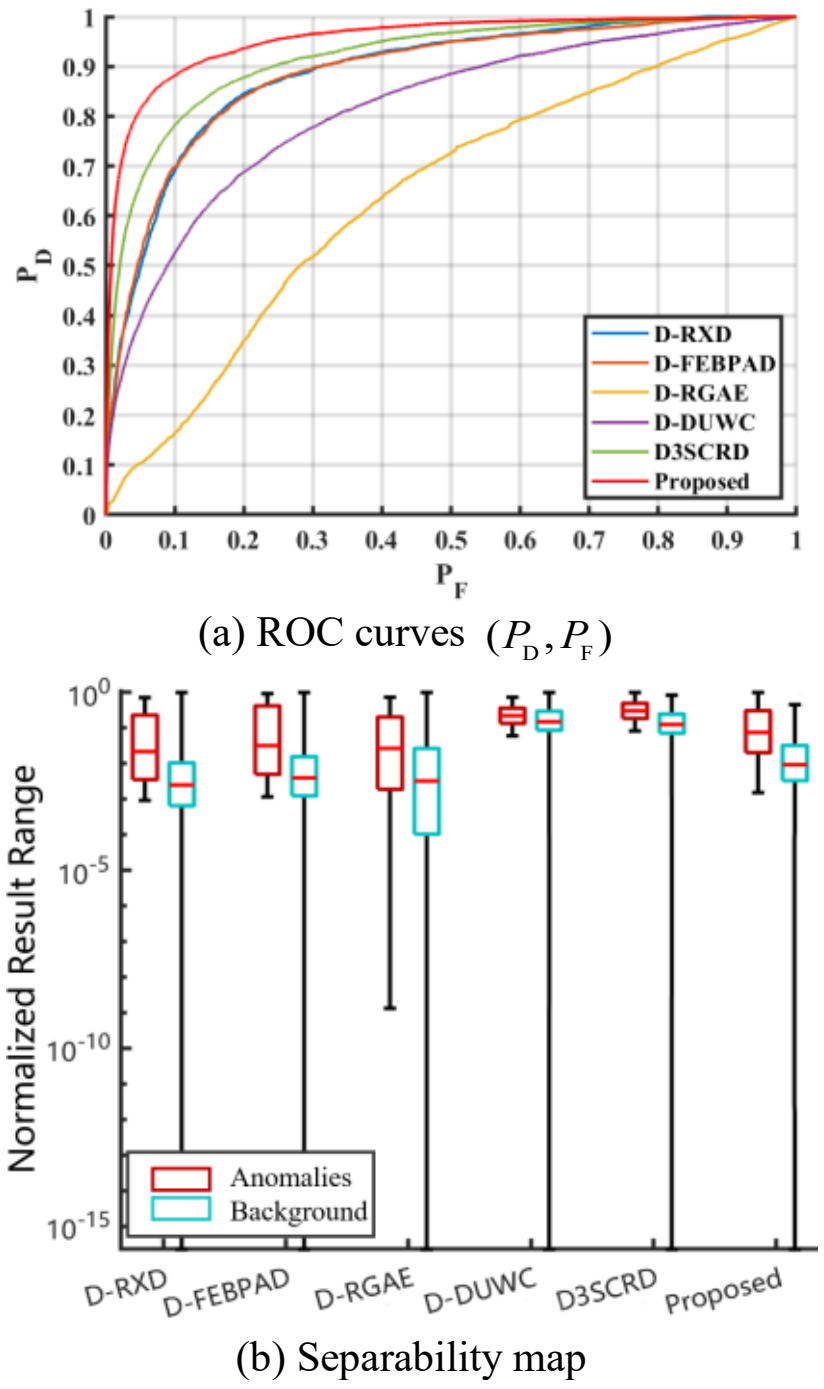

Fig. 8. ROC curves and separability map of results.

## IV. CONCLUSION

This study proposes an illumination-invariant anomaly detection framework in sub-canopy UAV multispectral point clouds. Addressing the limitations of conventional methods that rely on external metadata or sequential compensation, our prior-free shadow extraction strategy innovatively formulates solar angle estimation as an inverse optimization problem. Furthermore, the proposed illumination-consistent sparse representation effectively disentangles intrinsic material reflectance from extrinsic lighting variations. By enforcing strict physical consistency during dictionary construction, this approach significantly improves the separability of anomalies in unevenly illuminated environments, providing a practical solution for forest anomaly detection. Despite its effectiveness, the performance may be constrained under conditions of exceptionally dense vegetation or rapidly shifting illumination. Future research will focus on enhancing the algorithm's computational efficiency and verifying its generalization across more diverse forest structures.

## ACKNOWLEDGMENT

This work was supported by National Natural Science Foundation of China through the Youth Science Foundation Project (Grant No. 62301192).

## REFERENCES

[1] L. Chen, Y. Gu, X. Li, X. Zhang, and B. Liu, “A Normalized Spatial–Spectral Supervoxel Segmentation Method for Multispectral Point Cloud Data,” IEEE Trans. Geosci. Remote Sens., vol. 61, pp. 1–11, 2023, doi: 10.1109/TGRS.2023.3313734.

[2] Y. Gu, C. Wang, and X. Li, “An Intensity-Independent Stereo Registration Method of Push-Broom Hyperspectral Scanner and LiDAR on UAV Platforms,” IEEE Trans. Geosci. Remote Sens., vol. 60, pp. 1–14, 2022, doi: 10.1109/TGRS.2022.3211202.

[3] Q. Guo, Y. Cen, L. Zhang, Y. Zhang, and Y. Huang, “Hyperspectral Anomaly Detection Based on Spatial–Spectral Cross-Guided Mask Autoencoder,” IEEE J. Sel. Top. Appl. Earth Obs. Remote Sens., vol. 17, pp. 9876–9889, 2024, doi: 10.1109/JSTARS.2024.3393995.

[4] L. Chen, Y. Gu, and X. Li, “A local anomaly detection method for occluded targets with multispectral point clouds,” in IGARSS 2025 -

2025 IEEE International Geoscience and Remote Sensing Symposium, Brisbane, Australia: IEEE, Aug. 2025, pp. 7074–7077. doi: 10.1109/IGARSS55030.2025.11243660.

[5] S. Liu, L. Zhang, Y. Cen, L. Chen, and Y. Wang, “A Fast Hyperspectral Anomaly Detection Algorithm Based on Greedy Bilateral Smoothing and Extended Multi-Attribute Profile,” Remote Sens., vol. 13, no. 19, p. 3954, Oct. 2021, doi: 10.3390/rs13193954.

[6] T. Zhou, H. Fu, C. Sun, and S. Wang, “Shadow Detection and Compensation from Remote Sensing Images under Complex Urban Conditions,” Remote Sens., vol. 13, no. 4, p. 699, Feb. 2021, doi: 10.3390/rs13040699.

[7] X. Dong, J. Cao, and W. Zhao, “A review of research on remote sensing images shadow detection and application to building extraction,” Eur. J. Remote Sens., vol. 57, no. 1, p. 2293163, Dec. 2024, doi: 10.1080/22797254.2023.2293163.

[8] I. S. Reed and X. Yu, “Adaptive multiple-band CFAR detection of an optical pattern with unknown spectral distribution,” IEEE Trans. Acoust. Speech Signal Process., vol. 38, no. 10, pp. 1760–1770, Oct. 1990, doi: 10.1109/29.60107.

[9] Y. Ma, G. Fan, Q. Jin, J. Huang, X. Mei, and J. Ma, “Hyperspectral Anomaly Detection via Integration of Feature Extraction and Background Purification,” IEEE Geosci. Remote Sens. Lett., vol. 18, no. 8, pp. 1436–1440, Aug. 2021, doi: 10.1109/LGRS.2020.2998809.

[10] G. Fan, Y. Ma, X. Mei, F. Fan, J. Huang, and J. Ma, “Hyperspectral Anomaly Detection With Robust Graph Autoencoders,” IEEE Trans. Geosci. Remote Sens., vol. 60, pp. 1–14, 2022, doi: 10.1109/TGRS.2021.3097097.

[11] H. Zhang, Z. Liu, C. Jiao, B. Niu, and F. Li, “Discriminative spectral analysis via dual-window coupled modeling for hyperspectral anomaly target detection,” IEEE Sens. J., vol. 25, no. 16, pp. 30998–31009, Aug. 2025, doi: 10.1109/JSEN.2025.3582158.

[12] L. Chen, Y. Gu, and X. Li, “Unsupervised occluded target detection based on spherical shell with multispectral point clouds,” IEEE Trans. Geosci. Remote Sens., vol. 63, pp. 1–13, 2025, doi: 10.1109/TGRS.2025.3585524.